\def\year{2022}\relax
\documentclass[letterpaper]{article} 
\usepackage{aaai22}  
\usepackage{times}  
\usepackage{helvet}  
\usepackage{courier}  
\usepackage[hyphens]{url}  
\usepackage{graphicx} 
\usepackage{natbib}  
\usepackage{caption} 
\DeclareCaptionStyle{ruled}{labelfont=normalfont,labelsep=colon,strut=off} 
\usepackage{amsmath}
\usepackage{amssymb}

\usepackage{colortbl}
\usepackage{arydshln}
\usepackage{multirow}
\usepackage{multicol}
\usepackage{dashrule}
\usepackage{fancyhdr}
\usepackage{bm}

\usepackage{algorithm}
\usepackage{algorithmic}
\usepackage{newfloat}
\usepackage{listings}
\usepackage[pagebackref=false,breaklinks=false,colorlinks,bookmarks=false]{hyperref}
\floatstyle{ruled}
\newfloat{listing}{tb}{lst}{}
\floatname{listing}{Listing}
\title{TransZero: Attribute-guided Transformer for Zero-Shot Learning}
\author{
	    Shiming Chen\textsuperscript{\rm 1}\equalcontrib,
	    Ziming Hong\textsuperscript{\rm 1}\equalcontrib,
	    Yang Liu\textsuperscript{\rm 2},
	    Guo-Sen Xie\textsuperscript{\rm 3},
	    Baigui Sun\textsuperscript{\rm 2},\\
	    Hao Li\textsuperscript{\rm 2},
	    Qinmu Peng\textsuperscript{\rm 1},
	    Ke Lu\textsuperscript{\rm 4},
	    Xinge You\textsuperscript{\rm 1\thanks{Corresponding author}}
}
\affiliations{
	 \textsuperscript{\rm 1}Huazhong University of Science and Technology (HUST), China\hspace{8mm}
	 \textsuperscript{\rm 2}Alibaba Group, Hangzhou, China\\
	 \textsuperscript{\rm 3}Mohamed bin Zayed University of AI (MBZUAI), UAE\hspace{8mm}
	 \textsuperscript{\rm 4} University of Chinese Academy of Sciences, China\\\vspace{2mm}
	 \{shimingchen, pengqinmu, youxg\}@hust.edu.cn \hspace{5mm} \{hoongzm, gsxiehm\}@gmail.com
	
}

\usepackage{bibentry}

\usepackage[switch]{lineno}
\begin{document}
	\maketitle
	
	\begin{abstract}
		Zero-shot learning (ZSL) aims to recognize novel classes by transferring semantic knowledge from seen classes to unseen ones. Semantic knowledge is learned from attribute descriptions shared between different classes, which act as strong priors for localizing object attributes that represent discriminative region features, enabling significant visual-semantic interaction. Although some attention-based models have attempted to learn such region features in a single image, the transferability and discriminative attribute localization of visual features are typically neglected. In this paper, we propose an attribute-guided Transformer network, termed TransZero, to refine visual features and learn attribute localization for discriminative visual embedding representations in ZSL. Specifically, TransZero takes a feature augmentation encoder to alleviate the cross-dataset bias between ImageNet and ZSL benchmarks, and improves the transferability of visual features by reducing the entangled relative geometry relationships among region features. To learn locality-augmented visual features, TransZero employs a visual-semantic decoder to localize the image regions most relevant to each attribute in a given image, under the guidance of semantic attribute information. Then, the locality-augmented visual features and semantic vectors are used to conduct effective visual-semantic interaction in a visual-semantic embedding network. Extensive experiments show that TransZero achieves the new state of the art on three ZSL benchmarks. The codes are available at: \url{https://github.com/shiming-chen/TransZero}.
	\end{abstract}

	\section{Introduction}\label{sec1}
	\noindent Inspired by human cognitive competence, zero-shot learning (ZSL) was proposed to recognize new classes during learning by exploiting the intrinsic semantic relatedness between seen and unseen classes \cite{Larochelle2008ZerodataLO,Palatucci2009ZeroshotLW,Lampert2009LearningTD}. In ZSL, there are no training samples available for unseen classes in the test set, and the label spaces for the training set and test set are disjoint from each other. Thus, the key task for ZSL is to learn discriminative visual features for conducting effective visual-semantic interactions based on the semantic information (e.g., sentence embeddings \cite{Reed2016LearningDR}, and attribute vectors \cite{Lampert2014AttributeBasedCF}), which are shared between the seen and unseen classes employed to support the knowledge transfer. According to their classification range, ZSL methods can be categorized into conventional ZSL (CZSL), which aims to predict unseen classes, and generalized ZSL (GZSL), which can predict both seen and unseen classes \cite{Xian2017ZeroShotLC}.
	
	\begin{figure}[t]
		\centering
		\includegraphics[scale=0.34]{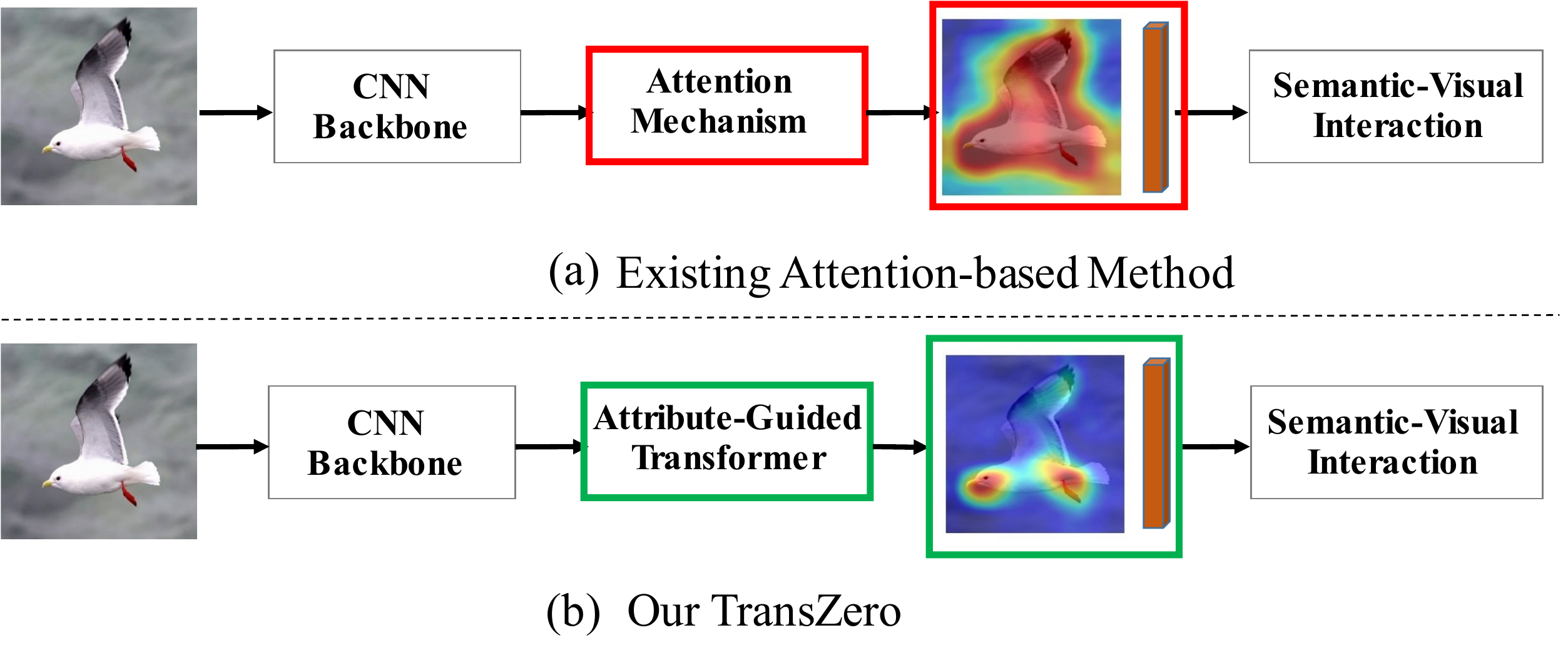}\vspace{-2mm}
		\caption{Motivation illustration. (a) Existing attention-based ZSL methods simply learn region embeddings (e.g., the whole bird body), neglecting the transferability and discriminative attribute localization (e.g., the distinctive bird body parts) of visual features; (b) Our TransZero reduces the entangled relationships among region features to improve their transferability and localizes the object attributes to represent discriminative region features, enabling significant visual-semantic interaction.}
		\label{fig:motivation}
	\end{figure}
	
	To enable visual-semantic interactions, early ZSL methods attempt to build an embedding between seen classes and their class semantic vectors, and then classify unseen classes by nearest neighbor search in the embedding space. Since the embedding is only learned by seen class samples, these embedding-based methods usually overfit to seen classes under the GZSL setting (known as the bias problem). To tackle this problem, many generative ZSL methods have been proposed to generate samples of unseen classes by leveraging generative models (e.g., variational autoencoders (VAEs) \cite{Arora2018GeneralizedZL,Schnfeld2019GeneralizedZA,Chen2021HSVA}, generative adversarial nets (GANs) \cite{Xian2018FeatureGN,Xian2019FVAEGAND2AF,Chen2021FREE}, and generative flows \cite{Shen2020InvertibleZR}) for data augmentation. Thus the ZSL task is converted into a supervised classification problem.
	
	Although these methods have achieved progressive improvement, they rely on global visual features which are insufficient for representing the fine-grained information of classes (e.g., red-legged of Kittiwake), since the discriminative information is contained in a few regions corresponding to a few attributes. Thus, the visual feature representations are limited, resulting in poor visual-semantic interactions. More recently, some attention-based models \cite{Xie2019AttentiveRE,Xie2020RegionGE,Zhu2019SemanticGuidedML,Xu2020AttributePN,Yu2018StackedSA,Liu2019AttributeAF} have attempted to learn more discriminative region features with the guidance of the semantic information, as shown in Fig. \ref{fig:motivation}(a). However, these methods are limited in: i) They directly take the entangled region (grid) features for ZSL classification, which hinders the transferability of visual features from seen to unseen classes; ii) They simply learn region embeddings (e.g., the whole bird body), neglecting the importance of discriminative attribute localization (e.g., the distinctive bird body parts). Thus, properly improving the transferability and localizing the object attributes for enabling significant visual-semantic interaction in ZSL has become very necessary.
	
	To tackle the above challenges, in this paper, we propose an attribute-guided Transformer, termed TransZero, which reduces the entangled relationships among region features to improve their transferability and localizes the object attributes to represent discriminative region features in ZSL, as shown in Fig. \ref{fig:motivation}(b). Specifically, TransZero consists of an attributed-guided Transformer (AGT) that learns locality-augmented visual features and a visual-semantic embedding network (VSEN) that conducts visual-semantic interactions. In AGT, we first take a feature augmentation encoder to i) alleviate the cross-dataset bias between ImageNet and ZSL benchmarks, and ii) reduce the entangled relative geometry relationships between different regions for improving the transferability from seen to unseen classes. They are ignored by existing ZSL methods. To learn locality-augmented visual features, we employ a visual-semantic detector in AGT to localize the image regions most relevant to each attribute in a given image, under the guidance of semantic attribute information. Then, the locality-augmented visual features and semantic vectors are used to enable visual-semantic interaction in VSEN.  Extensive experiments show that TransZero achieves the new state of the art on three ZSL benchmarks. The qualitative results also demonstrate that TransZero refines visual features and provides attribute-level localization. 
	
	The main contributions of this paper are summarized as follows:
	
	\begin{itemize}
		\item We introduce a novel ZSL method, termed TransZero, which employs an attribute-guided Transformer to refine the visual features and learn the attribute localization for discriminative visual embedding representations. To the best of our knowledge, TransZero is the first work extending the Transformer to the ZSL task.
		
		\item We propose a feature augmentation encoder to i) alleviate the cross-dataset bias between ImageNet and ZSL benchmarks, and ii) reduce the entangled relative geometry relationships between different regions to improve the transferability from seen to unseen classes. They are ignored by existing ZSL methods. 
		
		\item  Extensive experiments demonstrate that TransZero achieves the new state of the art on three ZSL benchmarks. We further qualitatively show that our TransZero refines the visual features and accurately localizes fine-grained parts for discriminative feature representations.
	\end{itemize}

	\begin{figure*}[ht]
		\centering
		\includegraphics[scale=0.34]{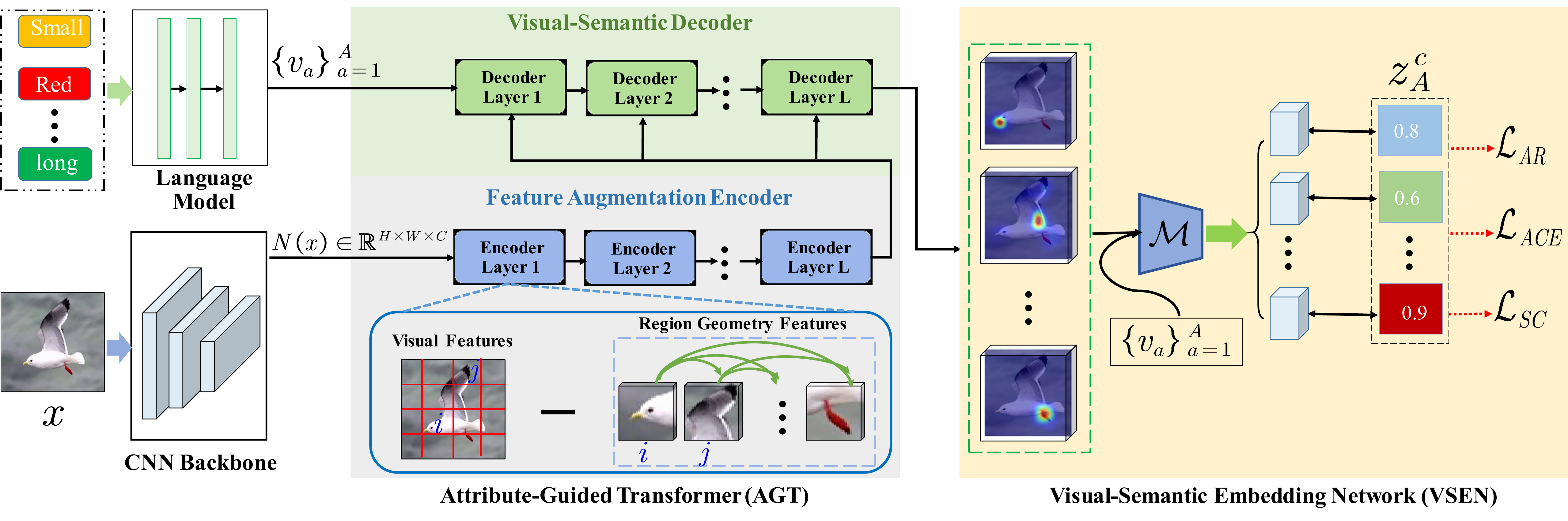}\vspace{-3mm}
		\caption{The architecture of the proposed TransZero model. TransZero consists of an attribute-guided Transformer (AGT) and a visual-semantic embedding network (VSEN). AGT includes a feature augmentation encoder that alleviates the cross-dataset bias between ImageNet and ZSL benchmarks and reduces the entangled geometry relationships between different regions for improving the transferability from seen to unseen classes, and a visual-semantic decoder that learns locality-augmented visual features based on the semantic attribute information. VSEN is used to enable significant visual-semantic interaction.}
		\label{fig:pipeline}\vspace{-6mm}
	\end{figure*}
	\section{Related Work}\label{sec2}
	\subsubsection{Zero-Shot Learning.}\label{sec2.1} Early ZSL methods \cite{Song2018TransductiveUE,Li2018DiscriminativeLO,Xian2018FeatureGN,Xian2019FVAEGAND2AF,Yu2020EpisodeBasedPG,Min2020DomainAwareVB,Han2021ContrastiveEF,Chen2021FREE,Chou2021AdaptiveAG, Han2021ContrastiveEF} focus on learning a mapping between the visual and semantic domains to transfer semantic knowledge from seen to unseen classes. They typically extract global visual features from pre-trained or end-to-end trainable networks. Typically, end-to-end models achieve better performance than pre-trained ones because they fine-tune the visual features, thus alleviating the cross-dataset bias between ImageNet and ZSL benchmarks \cite{Chen2021FREE,Xian2019FVAEGAND2AF}.  However, these methods still usually yield relatively undesirable results, since they cannot efficiently capture the subtle differences between seen and unseen classes. More relevant to this work are the recent attention-based ZSL methods \cite{Xie2019AttentiveRE,Xie2020RegionGE,Zhu2019SemanticGuidedML,Xu2020AttributePN,Liu2021GoalOrientedGE} that utilize attribute descriptions as guidance to discover the more discriminative region (or part) features. Unfortunately, They simply learn region embeddings (e.g., the whole bird body) neglecting the importance of discriminative attribute localization (e.g., the distinctive bird body parts). Furthermore, the end-to-end attention models are also time-consuming when it comes to fine-tuning the CNN backbone. In contrast, we propose an attribute-guided Transformer to learn the attribute localization for discriminative region feature representations under non end-to-end ZSL model.
	
	\subsubsection{Transformer Model.}\label{sec2.2} Transformer models \cite{Vaswani2017AttentionIA} have recently demonstrated excellent performance on a broad range of language and computer vision tasks, e.g., machine translation \cite{Ott2018ScalingNM}, image recognition \cite{Dosovitskiy2021AnII}, video understanding \cite{Gabeur2020MultimodalTF}, visual question answering \cite{ZhangRSTNetCW}, etc. The success of Transformers can be mainly attributed to self-supervision and self-attention \cite{Khan2021TransformersIV}. The self-supervision allows complex models to be trained without the high cost of manual annotation, which in turn enables generalizable representations that encode useful relationships between the entities presented in a given dataset to be learned. The self-attention layers take the broad context of a given sequence into account by learning the relationships between the elements in the token set  (e.g., words in language or patches in an image). Some methods \cite{Gabeur2020MultimodalTF,Cornia2020MeshedMemoryTF,Huang2019AttentionOA,Pan2020XLinearAN} have also shown that the transformer architecture can better capture the relationship between visual features and process sequences in parallel during training. Motivated by these, we design an attribute-guided Transformer that reduces the relationships among different regions to improve the transferability of visual features and learns the attribute localization for representing discriminative region features.

	\section{Proposed Method}\label{sec3}
	We first introduce some notations and the problem definition. Assume that we have $\mathcal{D}^{s}=\left\{\left(x_{i}^{s}, y_{i}^{s}\right)\right\}$ as training data with $C^s$ seen classes, where $x_i^s \in \mathcal{X}$ denotes the image $i$, and $y_i^s \in \mathcal{Y}^s$ is the corresponding class label. Another set of unseen classes $C^u$ has unlabeled samples $\mathcal{D}^{u}=\left\{\left(x_{i}^{u}, y_{i}^{u}\right)\right\}$, where $x_{i}^{u}\in \mathcal{X}$ are the unseen class images, and $y_{i}^{u} \in \mathcal{Y}^u$ are the corresponding labels. A set of class semantic vectors of the class $c \in \mathcal{C}^{s} \cup \mathcal{C}^{u} = \mathcal{C}$ with $A$ attributes $z^{c}=\left[z_{1}^{c}, \ldots, z_{A}^{c}\right]^{\top}= \phi(y)$ helps knowledge transfer from seen to unseen classes.  Note that we also use the semantic attribute vectors of each attribute $\left\{v_{a}\right\}_{a=1}^{A}$ learned by a language model (i.e., GloVe  \cite{Pennington2014GloveGV}) according to each word in attribute names. ZSL aims to predict the class labels $y^u \in \mathcal{Y}^u$ and $y \in \mathcal{Y}=\mathcal{Y}^{s} \cup \mathcal{Y}^{u}$ in the CZSL and GZSL settings, respectively, where $\mathcal{Y}^{s} \cap \mathcal{Y}^{u}=\emptyset$.
	
	In this paper, we propose an attribute-guided Transformer network (termed TransZero) to refine the visual features and localize the object attributes for representing the discriminative region features under a non end-to-end model. This enables significant visual-semantic interaction in ZSL. As illustrated in Fig. \ref{fig:pipeline}, our TransZero includes an attribute-guided Transformer (AGT) and visual-semantic embedding network (VSEN). AGT refines the visual feature using a feature augmentation encoder, and learns locality-augmented visual features using a visual-semantic decoder. VSEN enables visual-semantic interaction for ZSL classification.
	

	\subsection{Attribute-Guided Transformer}\label{sec3.2}
	\subsubsection{Feature Augmentation Encoder.}\label{sec3.2.1}
	Since there is a cross-dataset bias between ImageNet and ZSL benchmarks \cite{Chen2021FREE}, we introduce a feature augmentation encoder to refine the visual features of ZSL benchmarks. Additionally, previous ZSL methods typically flatten the grid features (extracted from a CNN backbone) into a feature vector, which is further used for generative models or embedding learning. However, such a feature vector implicitly entangles the feature representations among various regions in an image, which hinders their transferability from one domain to other domains (e.g., from seen to unseen classes) \cite{Xu2020AttributePN,Atzmon2020ACV,Chen2021CrossDomainFE}. As such, we propose a feature-augmented scaled dot-product attention to further enhance the encoder layer by reducing the relative geometry relationships among the grid features. 
	
	To learn relative geometry features \cite{Herdade2019ImageCT,ZhangRSTNetCW}, we first calculate the relative center coordinates ($v^{\text{cen}}_i$, $t^{\text{cen}}_i$) based on the pair of 2D relative positions of the $i$-th grid $\left\{\left(v_{i}^{\min }, t_{i}^{\min }\right),\left(v_{i}^{\max }, t_{i}^{\max }\right)\right\}$:
	\begin{align}
	\label{eq:center-coor}
	&\left(v^{\text{cen}}_{i}, t^{\text{cen}}_{i}\right)=\left(\frac{v_{i}^{\min }+v_{i}^{\max }}{2}, \frac{t_{i}^{\min }+t_{i}^{\max }}{2}\right),\\
	&w_{i}=\left(v_{i}^{\max }-v_{i}^{\min }\right)+1,\\
	&h_{i}=\left(t_{i}^{\max }-t_{i}^{\min }\right)+1,
	\end{align}
	where $(v_i^{\min }, t_i^{\min })$ and $(v_i^{\max }, t_i^{\max })$ are the relative position coordinates of the top left corner and bottom right corner of the grid $i$, respectively.
	
	Then, we construct region geometry features $G_{i j}$ between grid $i$ and grid $j$:
	\begin{gather}
	\label{eq:geometry-repre}
	G_{i j}=\operatorname{ReLU}\left(w_{g}^{T} g_{i j}\right),
	\end{gather}
	where
	\begin{gather}
	g_{i j}=F C\left(r_{i j}\right), \qquad r_{i j}=\left(\begin{array}{c}
	\log \left(\frac{\left|v^{\text{cen}}_{i}-v^{\text{cen}}_{j}\right|}{w_{i}}\right) \\
	\log \left(\frac{\left|t^{\text{cen}}_{i}-t^{\text{cen}}_{j}\right|}{h_{i}}\right) \\
	\end{array}\right),
	\end{gather}
	where $r_{ij}$ is the relative geometry relationship between grids $i$ and $j$, $FC$ is a fully connected layer followed by a $ReLU$ activation, and $w_{g}^{T}$ is a set of learnable weight parameters.
	
	Finally, we substract the region geometry features from the visual features in the feature-augmented scaled dot-product attention to provide a more accurate attention map, formally defined as: 
	\begin{align}
	\label{eq:encoder-atten}
	&Q^e=U W_{q}^e, K^e=U W_{k}^e, V^e=U W_{v}^e, \\
	&Z_{a u g}=\operatorname{softmax}\left(\frac{Q^e K^{e^{\top}}}{\sqrt{d^e}}-G\right) V^e,\\
	&U \leftarrow U+Z_{aug},
	\end{align}
	where $Q$, $K$, $V$ are the query, key and value matrices, $W_q^e$, $W_k^e$, $W_v^e$ are the learnable matrices of weights, $d^e$ is a scaling factor, and $Z_{aug}$ is the augmented features. $U\in \mathbb{R}^{HW \times C}$ are the packed visual features, which are learned from the flattened features embedded by a fully connected layer followed by a ReLU and a Dropout layer. 
	
	\subsubsection{Visual-Semantic Decoder.}\label{sec3.2.2}
	Following the standard Transformer \cite{Vaswani2017AttentionIA}, our visual-semantic decoder takes a multi-head self-attention layer and feed-forward network (FFN) to build the decoder layer. The decoding process continuously incorporates visual information under the guidance of semantic attribute features $v_A$. Thus, our visual-semantic decoder can effectively localize the image regions most relevant to each attribute in a given image. The multi-head self-attention layer uses the outputs of the encoder $U$ as keys ($K_t^d$) and values ($V_t^d$) and a set of learnable semantic embeddings $v_A$ as queries ($Q_t^d$). It is defined as:
	\begin{align}
	\label{eq:decoder-atten}
	&Q_t^d=v_A W_{qt}^d, K_t^d=U W_{kt}^d, V_t^d=U W_{vt}^d, \\
	&\text{head}_{t}=\operatorname{softmax}\left(\frac{Q_t^d K_t^{d^{\top}}}{\sqrt{d^d}}\right) V_t^d,\\
	&\hat{F}= \|_{t=1}^{T} (\text{head}_{t})W_o,
	\end{align}
	where $W_{qt}^d$, $W_{kt}^d$, $W_{vt}^d$ are the learnable weights, $d^d$ is a scaling factor, and $\|$ is a concatenation operation. Then, an FFN with two linear transformations followed a ReLU activation in between is applied to the attended features $\hat{F}$:
	\begin{gather}
	\label{eq:FFN}
	F=ReLu \left(\hat{F} W_{1}+b_{1}\right) W_{2}+b_{2},
	\end{gather}
	where $W_{1}$, $W_{2}$, $b_1$ and $b_2$ are the weights and biases of the linear layers respectively, and $F$ are the locality-augmented visual features.
	
	\subsection{Visual-Semantic Embedding Network}\label{sec3.3}
	After generating locality-augmented visual features, we further map them into the semantic embedding space. To encourage the mapping to be more accurate, we take the semantic attribute vectors $\{v_{a}\}_{a=1}^A$ as support, based on a mapping function ($\mathcal{M}$). Specifically, $\mathcal{M}$ matches the locality-augmented visual features $F$ with the semantic attribute information $v_{A}$:
	\begin{gather}
	\label{Eq:vs_encoder}
	\psi(x_i)=\mathcal{M}(F)= v_{A}^{\top} W F,
	\end{gather}
	where $W$ is an embedding matrix that embeds $F$ into the semantic attribute space. In essence, $\psi(x_i)[a]$ is an attribute score that represents the confidence of having the $a$-th attribute in the image $x_i$. Given a set of semantic attribute vectors $\left\{v_{a}\right\}_{a=1}^{A}$, TransZero attains a mapped semantic embedding $\psi(x_i)$.

	\subsection{Model Optimization}\label{sec3.4}
	To achieve effective optimization, we employ the attribute regression loss, attribute-based cross-entropy loss and self-calibration loss to train TransZero. 
	
	\subsubsection{Attribute Regression Loss.}
	To encourage VSEN to accurately map visual features into their corresponding semantic embeddings, we introduce an attribute regression loss to constrain TransZero. Here, we regard visual-semantic mapping as a regression problem and minimize the mean square error between the ground truth attribute $z^{c}$ and the embedded attribute score $\psi(x_i^s)$ of a sample $x_i^s$:
	\begin{gather}
	\label{eq:reg-loss}
	\mathcal{L}_{AR}=\|\psi(x_i^s)-z^{c}\|_{2}^{2}.
	\end{gather}
	
	\subsubsection{Attribute-Based Cross-Entropy Loss.}
	Since the associated image embedding is projected near its class semantic vector $z^{c}$ when an attribute is visually present in an image, we take the attribute-based cross-entropy loss $\mathcal{L}_{ACE}$ to optimize the parameters of the TransZero model, i.e., the dot product between the visual embedding and each class semantic vector is calculated to produce class logits. This encourages the image to have the highest compatibility score with its corresponding class semantic vector. Given a batch of $n_b$ training images $\{x_i^s\}_{i=1}^{n_b}$ with their corresponding class semantic vectors $z^c$,  $\mathcal{L}_{ACE}$ is defined as:
	\begin{gather}
	\label{eq:L_ACE}
	\mathcal{L}_{ACE}=-\frac{1}{n_{b}} \sum_{i=1}^{n_{b}} \log \frac{\exp \left(\psi(x_i^s) \times z^{c}\right)}{\sum_{\hat{c} \in \mathcal{C}^s} \exp \left(\psi(x_i^s) \times z^{\hat{c}} \right)}.
	\end{gather}
	
	\subsubsection{Self-Calibration Loss.}
	Since $\mathcal{L}_{AR}$ and $\mathcal{L}_{ACE}$ optimize the model on seen classes, TransZero inevitably overfits to these classes, as also observed in \cite{Zhu2019SemanticGuidedML,Huynh2020FineGrainedGZ,Xu2020AttributePN}. To tackle this challenge, we further introduce a self-calibration loss $\mathcal{L}_{SC}$ to explicitly shift some of the prediction probabilities from seen to unseen classes. $\mathcal{L}_{SC}$ is thus formulated as:
	\begin{gather}
	\label{eq:L_SC}
	\mathcal{L}_{SC}=-\frac{1}{n_{b}} \sum_{i=1}^{n_{b}}   \sum_{c^{\prime=1}}^{\mathcal{C}^u} \log \frac{\exp \left(\psi(x_i^s) \times z^{c^{\prime}} + \mathbb {I}_{\left[c^{\prime}\in\mathcal{C}^u\right]}\right)}{\sum_{\hat{c} \in \mathcal{C}} \exp \left(\psi(x_i^s) \times z^{\hat{c}} + \mathbb {I}_{\left[\hat{c}\in\mathcal{C}^u\right]}\right)},
	\end{gather}
	where $\mathbb {I}_{\left[c\in \mathcal{C}^u\right]}$ is an indicator function (i.e., it is 1 when $c\in\mathcal{C}^u$, otherwise -1). Intuitively, $\mathcal{L}_{ACE}$ encourages non-zero probabilities to be assigned to the unseen classes during training, which allows TransZero to produce a (large) non-zero probability for the true unseen class when given test samples from unseen classes. 
	
	\begin{table*}[ht]
		\centering  
		\caption{Results ~($\%$) of the state-of-the-art CZSL and GZSL modes on CUB, SUN and AWA2, including end-to-end and non end-to-end methods (generative and non-generative methods). The best and second-best results are marked in \textbf{\color{red}Red} and \textbf{\color{blue}Blue}, respectively. The Symbol “--” indicates no results. The Symbol “{\color{red}*}” denotes attention-based methods. }\label{Table:SOTA}\vspace{-2mm}
		\resizebox{1.0\linewidth}{!}{
			\begin{tabular}{r||c|ccc||c|ccc||c|ccc}
				\hline
				\multirow{3}{*}{\textbf{Methods}} 
				&\multicolumn{4}{c|}{\textbf{CUB}}&\multicolumn{4}{c|}{\textbf{SUN}}&\multicolumn{4}{c}{\textbf{AWA2}}\\
				\cline{2-5}\cline{6-9}\cline{9-13}
				&\multicolumn{1}{c|}{CZSL}&\multicolumn{3}{c|}{GZSL}&\multicolumn{1}{c|}{CZSL}&\multicolumn{3}{c|}{GZSL}&\multicolumn{1}{c|}{CZSL}&\multicolumn{3}{c}{GZSL}\\
				\cline{2-5}\cline{6-9}\cline{9-13}
				\textbf{} 
				&\rm{acc}&\rm{U} & \rm{S} & \rm{H} &\rm{acc}&\rm{U} & \rm{S} & \rm{H} &\rm{acc}&\rm{U} & \rm{S} & \rm{H} \\
				\hline 
				\textbf{End-to-End} &&&&&&&&&&&&\\   
				QFSL~\cite{Song2018TransductiveUE}&58.8&33.3&48.1&39.4&56.2&30.9&18.5&23.1&63.5&52.1&72.8&60.7\\
				LDF~\cite{Li2018DiscriminativeLO} &67.5&26.4&\textbf{\color{red}81.6}&39.9&--&--&--&--&65.5&9.8& 87.4& 17.6 \\
				SGMA$^{\color{red}*}$~\cite{Zhu2019SemanticGuidedML} &71.0&36.7&71.3&48.5&--&--&--&--&68.8&37.6&87.1&52.5\\
				AREN$^{\color{red}*}$~\cite{Xie2019AttentiveRE}&71.8&38.9&78.7&52.1&60.6&19.0&38.8&25.5&67.9&15.6&\textbf{\color{blue}92.9}&26.7 \\
				LFGAA$^{\color{red}*}$~\cite{Liu2019AttributeAF}&67.6&36.2&\textbf{\color{blue}80.9}&50.0&61.5&18.5&\textbf{\color{blue}40.0}&25.3&68.1&27.0&\textbf{\color{red}93.4}&41.9\\
				APN$^{\color{red}*}$~\cite{Xu2020AttributePN}&72.0&\textbf{\color{blue}65.3}& 69.3&\textbf{\color{blue}67.2}&61.6& 41.9&34.0&37.6&68.4&57.1&72.4&63.9\\
				
				\hline
				\textbf{Non End-to-End} &&&&&&&&&&&&\\ 
				\textbf{\textit{Generative Methods}} &&&&&&&&&&&&\\ 
				f-CLSWGAN~\cite{Xian2018FeatureGN}    &57.3&43.7&57.7& 49.7&60.8&42.6&36.6&39.4&68.2&57.9&61.4&59.6\\
				f-VAEGAN-D2~\cite{Xian2019FVAEGAND2AF}&61.0&48.4&60.1& 53.6&\textbf{\color{blue}64.7}&45.1&38.0&41.3&71.1&57.6&70.6&63.5\\
				OCD-CVAE~\cite{Keshari2020GeneralizedZL}&--&44.8&59.9&51.3&--&44.8&{\color{red}\textbf{42.9}}&\textbf{\color{red}43.8}&--&59.5&73.4&65.7\\
				E-PGN~\cite{Yu2020EpisodeBasedPG}&\textbf{\color{blue}72.4}&52.0&61.1&56.2&--&--&--&--&\textbf{\color{red}73.4}&52.6&83.5&64.6\\
				Composer~\cite{Huynh2020CompositionalZL}&69.4&56.4&63.8&59.9&62.6& \textbf{\color{red}55.1}&22.0& 31.4&\textbf{\color{blue}71.5}& \textbf{\color{blue}62.1}&77.3&\textbf{\color{blue}68.8}\\
				GCM-CF~\cite{Yue2021CounterfactualZA}&--&61.0&59.7&60.3&--& 47.9&37.8& 42.2&--& 60.4&75.1&67.0\\
				FREE~\cite{Chen2021FREE}&--&55.7&59.9&57.7&--& 47.4&37.2& 41.7&--& 60.4&75.4&67.1\\
				HSVA~\cite{Chen2021HSVA}&62.8&52.7&58.3&55.3&63.8&48.6&39.0&\textbf{\color{blue}43.3}&--&59.3&76.6&66.8\\
				
				\cdashline{1-13}[4pt/1pt]
				\textbf{\textit{Non-Generative Methods}} &&&&&&&&&&&&\\ 
				SP-AEN~~\cite{Chen2018ZeroShotVR}      &55.4&34.7&70.6&46.6 &59.2&24.9&38.6&30.3&58.5&23.3&90.9&37.1 \\
				PQZSL~\cite{Li2019CompressingUI}     &--&43.2&51.4&46.9 &--&35.1& 35.3&35.2&--&31.7& 70.9& 43.8 \\
				IIR~~\cite{Cacheux2019ModelingIA}&63.8&30.4&65.8&41.2&63.5&22.0&34.1&26.7&67.9&17.6&87.0&28.9\\
				TCN~\cite{Jiang2019TransferableCN}   &59.5&52.6&52.0&52.3&61.5&31.2&37.3&34.0&71.2&61.2&65.8&63.4\\
				DVBE~\cite{Min2020DomainAwareVB}&--&53.2&60.2&56.5&--&45.0&37.2&40.7&--&\textbf{\color{red}63.6}&70.8&67.0\\ 
				DAZLE$^{\color{red}*}$~\cite{Huynh2020FineGrainedGZ}&66.0&56.7&59.6&58.1&59.4&52.3&24.3&33.2&67.9&60.3&75.7&67.1\\
				
				\cdashline{2-13}[4pt/1pt]
				{\textbf{TransZero}}{~\textbf{(Ours)}}    &\textbf{\color{red}76.8}&\textbf{\color{red}69.3}&68.3&\textbf{\color{red}{68.8}}&\textbf{\color{red}65.6}&\textbf{\color{blue}52.6}&33.4&40.8&70.1&61.3&82.3&\textbf{\color{red}70.2}\\
				\hline	
		\end{tabular} }
		\label{table:sota}\vspace{-6mm}
	\end{table*}
	
	Finally, we formulate the overall loss function of TransZero:
	\begin{gather}
	\label{Eq:L_final}
	\mathcal{L}_{total}=  \mathcal{L}_{ACE} + \lambda_{AR}\mathcal{L}_{AR}+ \lambda_{SC}\mathcal{L}_{SC},
	\end{gather}
	where $\lambda_{AR}$ and $\lambda_{SC}$ are the weights to controle their corresponding loss terms.

	\subsection{Zero-Shot Prediction}
	After training TransZero, we first obtain the embedding features of a test instance $x_i$ in the semantic space i.e., $\psi(x_i)$. Then, we take an explicit calibration to predict the test label of $x_i$, which is formulated as:
	\begin{gather}
	\label{Eq:prediction}
	c^{*}=\arg \max _{c \in \mathcal{C}^u/\mathcal{C}}\psi(x_i) \times z^{c}+\mathbb {I}_{\left[c\in\mathcal{C}^u\right]}.
	\end{gather}
	Here, $\mathcal{C}^u/\mathcal{C}$ corresponds to the CZSL/GZSL setting respectively.

	\section{Experiments}\label{sec4}
	\subsubsection{Dataset} 	
	Our extensive experiments are conducted on three popular ZSL benchmarks, including two fine-grained datasets (e.g., CUB \cite{Welinder2010CaltechUCSDB2} and SUN \cite{Patterson2012SUNAD}) and a coarse-grained dataset (e.g., AWA2 \cite{Xian2017ZeroShotLC}). CUB has 11,788 images of 200 bird classes (seen/unseen classes = 150/50) depicted with 312 attributes. SUN includes 14,340 images from 717 scene classes (seen/unseen classes = 645/72) depicted with 102 attributes. AWA2 consists of 37,322 images from 50 animal classes (seen/unseen classes = 40/10) depicted with 85 attributes.
	\subsubsection{Evaluation Protocols}
	Following \cite{Xian2017ZeroShotLC}, we measure the top-1 accuracy both in the CZSL and GZSL settings. In the CZSL setting, we simply predict the unseen classes to compute the accuracy of test samples, i.e., $\bm{acc}$. In the GZSL setting,  we compute the accuracy of the test samples from both the seen classes (denoted as $\bm{S}$) and unseen classes (denoted as $\bm{U}$). Meanwhile, their harmonic mean (defined as $\bm{H =(2 \times S \times U) /(S+U)}$) is also employed for evaluation in the GZSL setting.
	
	\subsubsection{Implementation Details}
	We use the training splits proposed in \cite{Xian2018FeatureGN}. We take a ResNet101 pre-trained on ImageNet as the CNN backbone for extracting the feature map without fine-tuning. We use the SGD optimizer with hyperparameters (momentum = 0.9, weight decay = 0.0001) to optimize our model. The learning rate and batch size are set to 0.0001 and 50, respectively. We empirically set $\lambda_{SC}$ to 0.3 and $\lambda_{AR}$ to 0.005 for all datasets. The encoder and decoder layers are set to 1 with one attention head.
	
	\begin{figure*}[t]
		\begin{center}
			\hspace{0.5mm}\rotatebox{90}{\hspace{0.4cm}{\footnotesize (a) AREN }}\hspace{-1mm}
			\includegraphics[width=17cm,height=2.05cm]{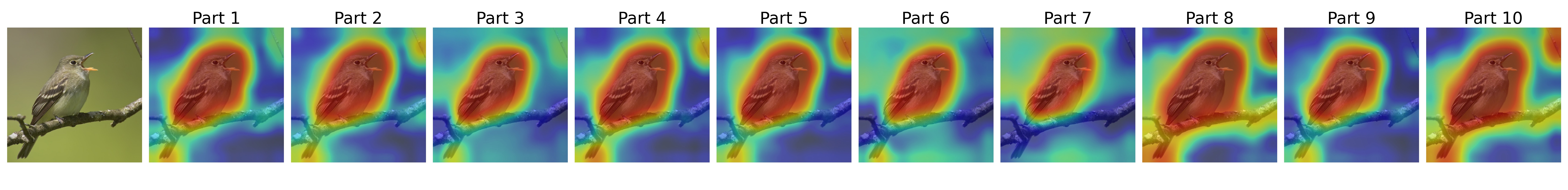}\\\vspace{-2mm}
			\hspace{0.5mm}\rotatebox{90}{\hspace{0.4cm}{\footnotesize (b) TransZero }}\hspace{-1mm}
			\includegraphics[width=17cm,height=2.55cm]{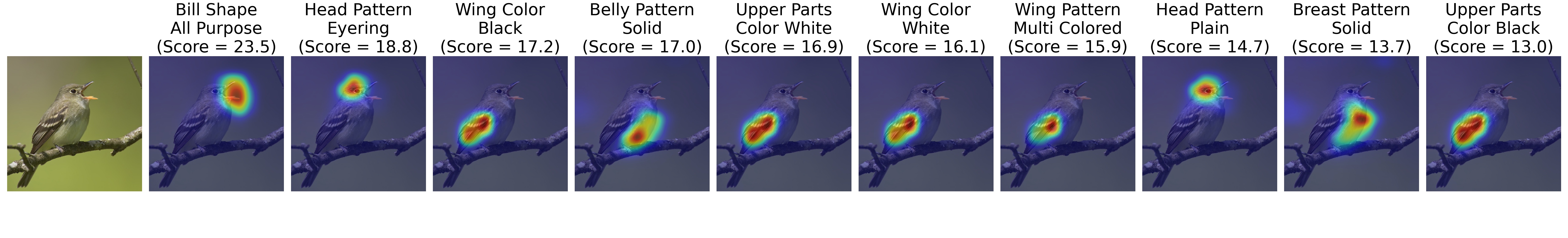}\\\vspace{-7mm}
			\caption{Visualization of attention maps for the attention-based method (i.e, AREN \cite{Xie2019AttentiveRE}) and our TransZero.}
			\label{fig:attened-part}\vspace{-6mm}
		\end{center}
	\end{figure*}

	\subsection{Comparison with State of the Art}\label{sec4.1}	
	
	\subsubsection{Conventional Zero-Shot Learning.} Here, we first compare our TransZero with the state-of-the-art methods in the CZSL setting. As shown in Table \ref{table:sota}, our TransZero achieves the best accuracies of 76.8\% and 65.6\% on CUB and SUN, respectively. This shows that TransZero effectively learns the attribute-augmented region feature representations for distinguishing various fine-grained classes.  As for the coarse-grained dataset (i.e., AWA2), TransZero still obtains competitive performance, with a top-1 accuracy of 70.1\%. Compared with other attention-based methods (e.g., SGMA~\cite{Zhu2019SemanticGuidedML}, AREN~\cite{Xie2019AttentiveRE}, APN~\cite{Xu2020AttributePN}), TransZero obtains significant gains of over 4.8\% and 4.0\% on CUB and SUN, respectively. This demonstrates that the attribute localization representations learned by our TransZero are more discriminative than the region embeddings learned by the existing attention-based methods on fine-grained datasets.

	\subsubsection{Generalized Zero-Shot Learning.}
	Table \ref{table:sota} shows the results of different methods in the GZSL setting. The results show that the unseen accuracy ($\bm{U}$) of all methods is usually lower than the seen accuracy ($\bm{S}$) on the CUB and AWA2 datasets, i.e., $\bm{U}<\bm{S}$. Meanwhile, $\bm{U}>\bm{S}$ on the SUN dataset since the number of seen classes is much larger than the number of unseen classes.
	
	We can see that most state-of-the-art methods achieve good results on seen classes but fail on unseen classes, while our method generalizes better to unseen classes with high unseen and seen accuracies.  For example, TransZero achieves the best performance with harmonic mean of 68.8\% and 70.2\% on CUB and AWA2, respectively. We argue that the benefits of TransZero come from the fact that i) the feature augmentation encoder in AGT improves the discriminability and transferability of visual features, and ii) the self-calibration mechanism alleviates the bias problem. Finally, our TransZero also outperforms the attention-based methods by harmonic mean improvements of at least 1.6\%, 3.2\% and 3.1\% on CUB, SUN and AWA2, respectively. This demonstrates the superiority and great potential of our attribute-guided Transformer for the ZSL task.

	\begin{table}[t]
		\centering
		\caption{ Ablation studies for different components of TransZero on the CUB and SUN datasets. “FAE” is the feature augmentation encoder, “FA” means feature augmentation, and “DEC” denotes visual-semantic decoder.} \label{table:ablation}
		\vspace{-2mm}
		\resizebox{0.49\textwidth}{!}
		{
			\begin{tabular}{l|c|ccc|c|ccc}
				\hline
				\multirow{2}*{Method} &\multicolumn{4}{c|}{CUB} &\multicolumn{4}{c}{SUN}\\
				\cline{2-5}\cline{6-9}
				&\rm{acc}&\rm{U} & \rm{S} & \rm{H} &\rm{acc}&\rm{U} & \rm{S} & \rm{H}\\
				\hline
				TransZero w/o FAE                                                  & 67.3& 61.0 &53.1 & 56.8& 61.2& 55.7&22.5&32.1\\
				TransZero w/o FA                                                   & 74.0& 66.7&66.3&66.5& 63.8& 49.5&31.4&38.5\\
				TransZero w/o DEC                                    & 62.3& 53.3&54.1&53.7& 58.3& 35.0&28.8&31.6\\
				TransZero w/o $\mathcal{L}_{SC}$                                   & 74.8&47.1&75.5&58.1& 64.2& 42.4&33.4&37.4\\
				TransZero w/o $\mathcal{L}_{AR}$                                   & 74.5& 65.9&68.8&67.3& 64.1& 47.2&33.3&39.1\\
				TransZero (full)                                                   & \textbf{76.8}&69.3&68.3&\textbf{68.8}&\textbf{65.6}&52.6&33.4&\textbf{40.8}\\
				\hline
			\end{tabular}
		}\vspace{-5mm}
	\end{table}
	
	\subsection{Ablation Study}\label{sec4.2}	
	To provide the further insight into TransZero, we conduct ablation studies to evaluate the effects of the feature augmentation encoder (denoted as FAE), feature augmentation in FAE (denoted as FA), visual-semantic decoder (denoted as DEC), self-calibration loss (i.e., $\mathcal{L}_{SC}$) and attribute regression loss (i.e.,  $\mathcal{L}_{AR}$). Our results are shown in Table \ref{table:ablation}. TransZero performs significantly worse than its full model when no feature augmentation encoder is used, i.e., the acc/harmonic mean drops by 9.5\%/12.0\% on CUB and 4.4\%/8.7\% on SUN. If we incorporate the encoder of the standard Transformer without feature augmentation, TransZero again achieves poor results compared to its full model, i.e., the acc/harmonic mean drops by 2.8\%/2.3\% and 1.8\%/2.3\% on CUB and SUN, respectively. When TransZero without visual-semantic decoder, its performances decreases dramatically on all datasets. Moreover, the self-calibration mechanism can effectively alleviate the bias problem, resulting in improvements in the harmonic mean of 10.7\% and 3.4\% on CUB and SUN, respectively. The attribute regression constraint further improves the performance of TransZero by directing VSEN to conduct effective visual-semantic mapping.

	\subsection{Qualitative Results}\label{sec4.3}	
	\subsubsection{Visualization of Attention Maps.} To intuitively show the effectiveness of our TransZero at learning locality-augmented visual features, we visualize the attention maps learned by the existing attention-based methods (e.g., AREN \cite{Xie2019AttentiveRE}) and TransZero. As shown in Fig. \ref{fig:attened-part}, AREN simply learns region embeddings for visual representations, e.g., the whole bird body, neglecting the fine-grained semantic attribute information. In contrast, our Transzero learns discriminative attribute localization for visual features by assigning high positive scores to key attributes (e.g., the ‘\textit{bill shape all purpose}’ of the Acadian Flycatcher in Fig. \ref{fig:attened-part}). Thus, TransZero achieves significant performance both in seen and unseen classes.

	\begin{figure*}[t]
		\begin{center}
			\hspace{0.5mm}\rotatebox{90}{\hspace{0.7cm}{\footnotesize (a) Seen Classes }}\hspace{0mm}
			\includegraphics[width=16cm,height=4cm]{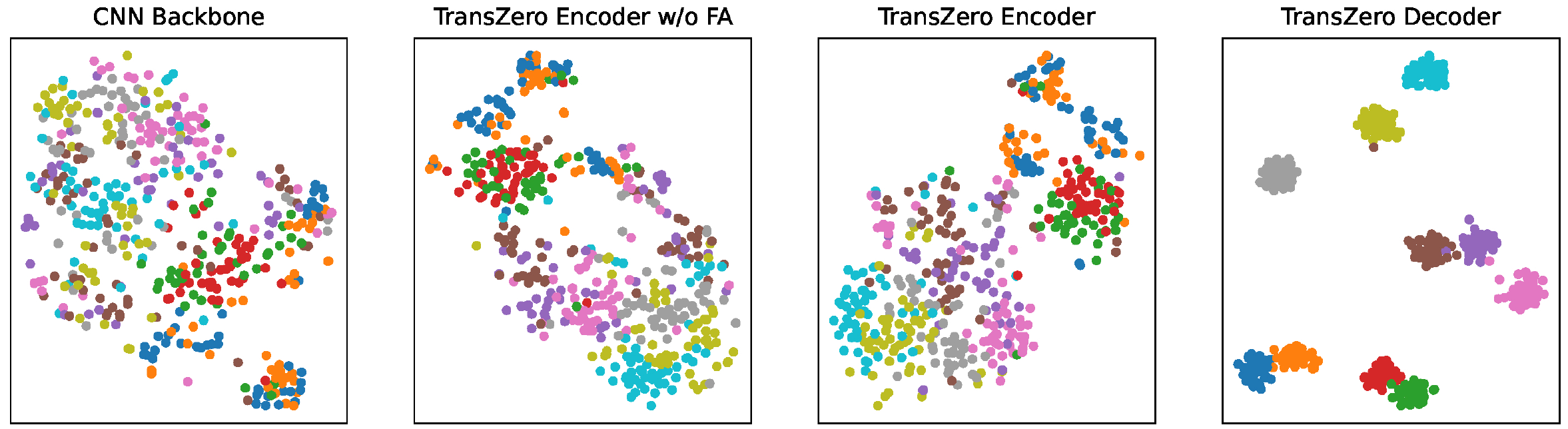}\\ 
			\hspace{0.5mm}\rotatebox{90}{\hspace{0.5cm}{\footnotesize (b) Unseen Classes }}\hspace{0.7mm}
			\includegraphics[width=16cm,height=4cm]{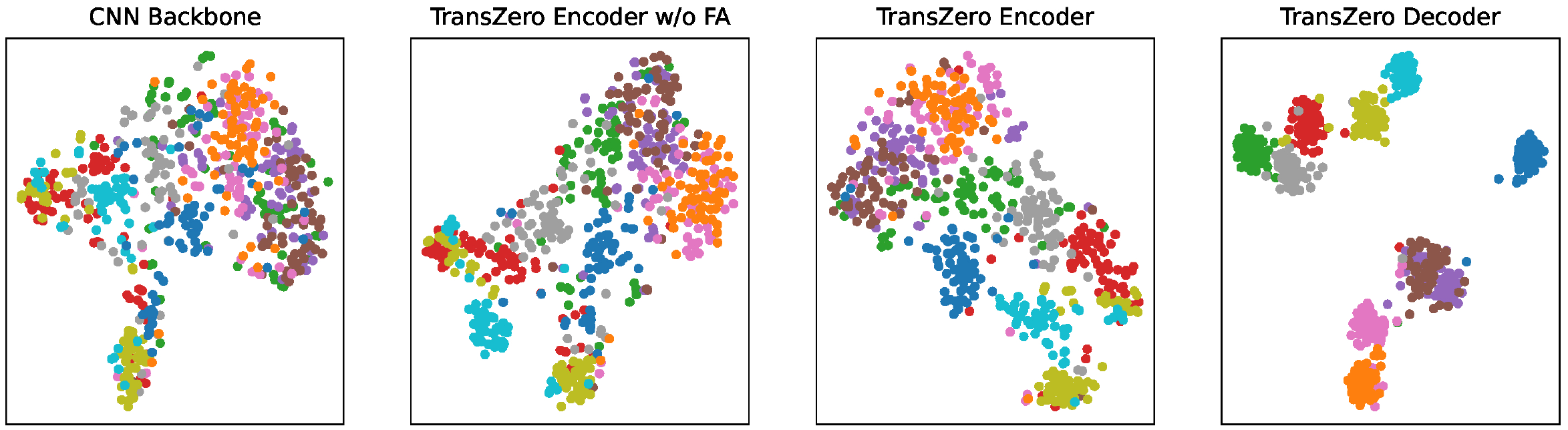}\\\vspace{-2mm}
			\caption{t-SNE visualizations of visual features for (a) seen classes and (b) unseen classes, learned by the CNN backbone, TransZero encoder w/o FA, TransZero encoder, and TransZero decoder. The 10 colors denote 10 different seen/unseen classes randomly selected from CUB.}
			\label{fig:tsne}\vspace{-6mm}
		\end{center}
	\end{figure*}
	
	\begin{figure}[t]
		\begin{center}
			\includegraphics[width=4.1cm,height=3.2cm]{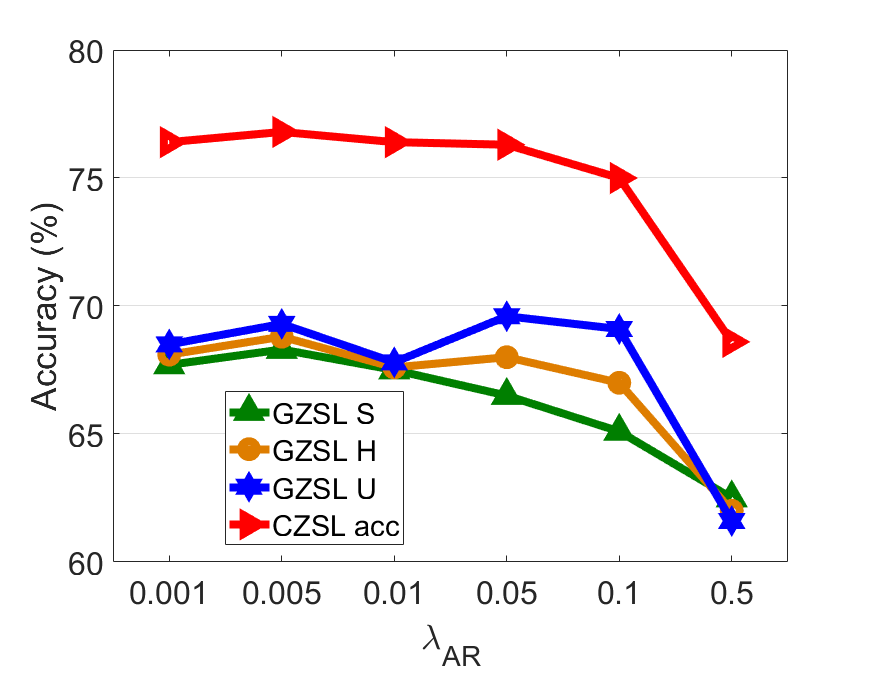}
			\includegraphics[width=4.1cm,height=3.2cm]{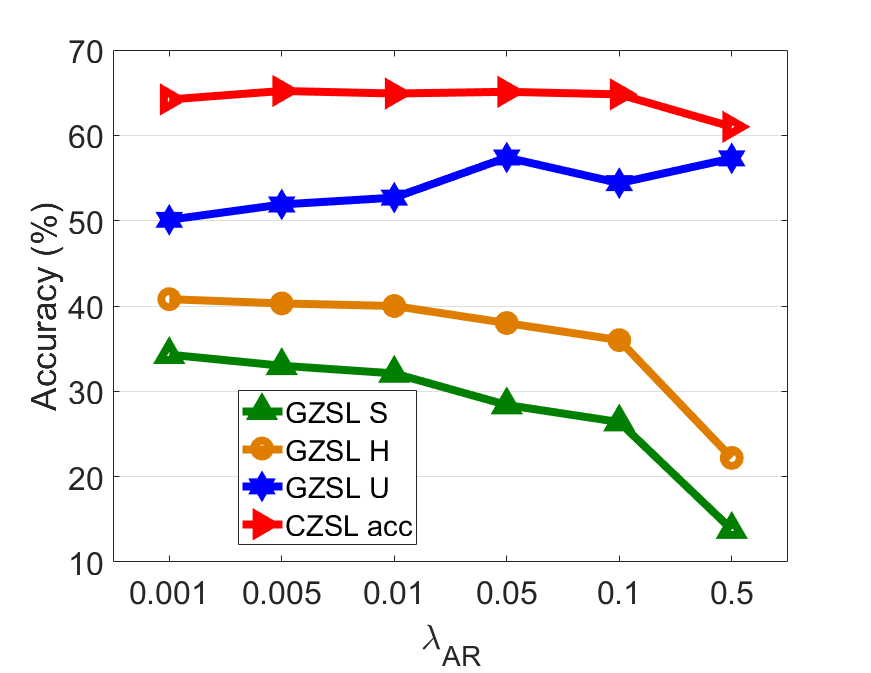}\\
			(a) CUB \hspace{3cm} (b) SUN
			\caption{The effects of $\lambda_{AR}$.}
			\label{fig:loss-AR}\vspace{-6mm}
		\end{center}
	\end{figure}
	
	\subsubsection{t-SNE Visualizations.} As shown in Fig. \ref{fig:tsne}, we present the t-SNE visualization \cite{Maaten2008VisualizingDU} of visual features for (a) seen classes and (b) unseen classes on CUB, learned by the CNN backbone, TransZero encoder w/o FA, TransZero encoder, and TransZero decoder. When we incorporate the standard encoder into our TransZero, the visual features learned by the encoder are significantly improved compared to the original visual features extracted from the CNN Backbone (e.g., ResNet101). When we use the feature augmentation encoder to refine the original visual features, the quality of the unseen features is further enhanced. These results demonstrate that the encoder of TransZero effectively alleviates the cross-dataset bias problem and reduces the entangled relative geometry relationships among different regions, improving the transferability. Moreover, the visual-semantic decoder learns locality-augmented visual features for improving visual feature representations.

	\subsection{Hyperparameter Analysis}\label{sec4.4}	
	\subsubsection{Effects of Loss Weight.} $\lambda_{AR}$ is employed to weigh the importance of the attribute regression loss, which directs the VSEN to conduct effective visual-semantic interaction. We try a wide range of $\lambda_{AR}$ evaluated on CUB and SUN, i.e., $\lambda_{AR}=\{0.001,0.005,0.01,0.05,0.1,0.5\}$. Results are shown in Fig. \ref{fig:loss-AR}. When $\lambda_{AR}$ is set to a large value, all evaluation protocols tend to drop. This is because the loss value of the attribute regression loss is too large, and thus the contributions of other losses are mitigated. When $\lambda_{AR}$ is set to 0.005, our TransZero achieves the best performance.
	
	\subsubsection{Effects of Loss Weight.} 
\begin{figure}[t]
		\begin{center}
			\includegraphics[width=4.1cm,height=3.2cm]{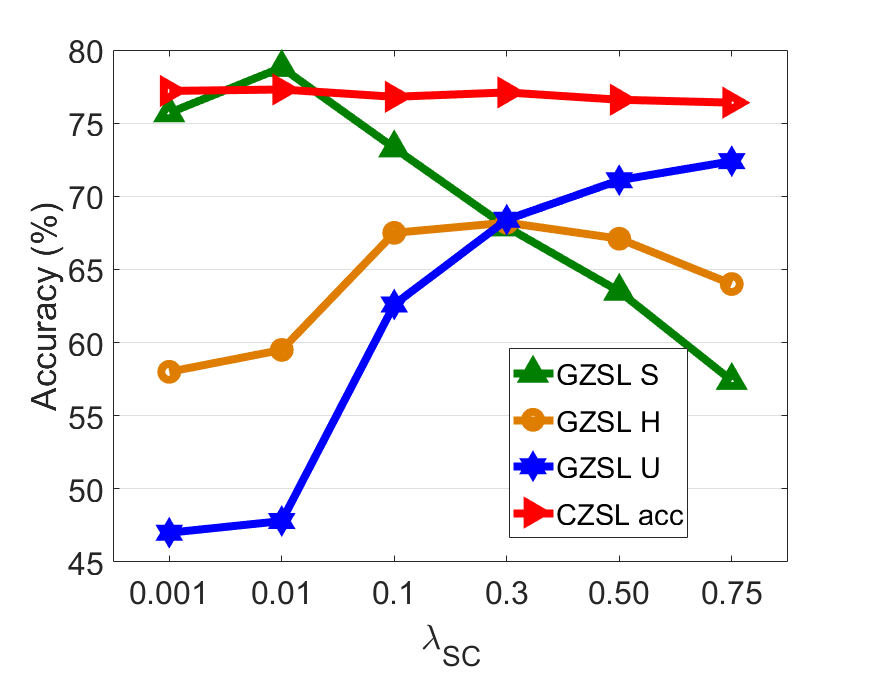}
			\includegraphics[width=4.1cm,height=3.2cm]{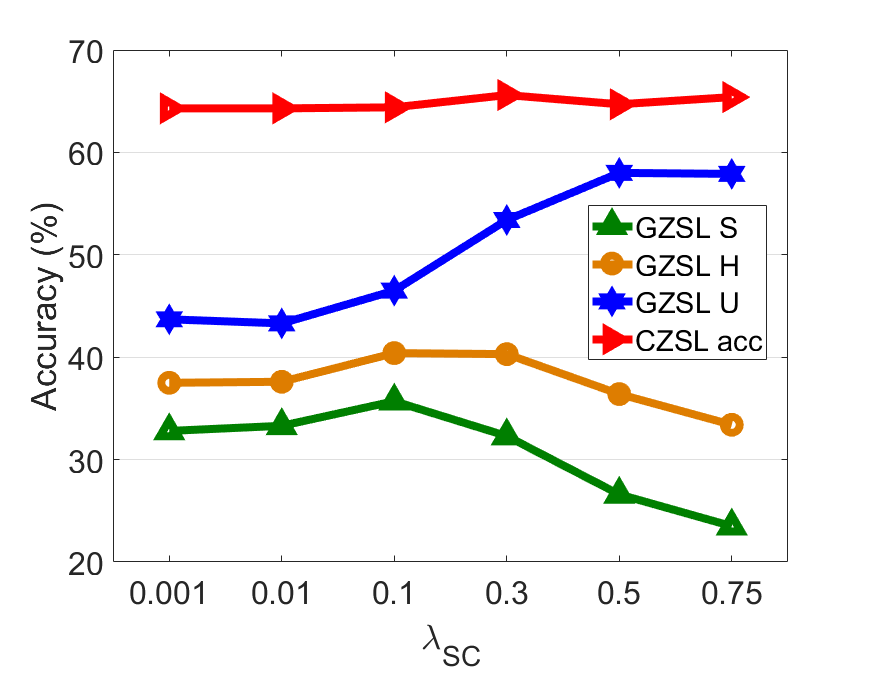}\\
			(a) CUB \hspace{3cm} (b) SUN 
			\caption{The effects of $\lambda_{SC}$.}
			\label{fig:loss-SC}\vspace{-6mm}
		\end{center}
	\end{figure}

$\lambda_{SC}$ adjusts the weight of the self-calibration loss, which effectively alleviates the bias problem. As shown in Fig. \ref{fig:loss-SC}, the accuracy on seen classes increases and the accuracy on unseen classes decreases when we increase $\lambda_{SC}$. Meanwhile, TransZero is insensitive to the self-calibration loss in the CZSL setting. We investigate a wide range of $\lambda_{SC}$ on CUB and SUN to find an appropriate setting for $\lambda_{SC}$.  Based on the results, we set $\lambda_{SC}$ to 0.3 for all datasets.

	\section{Conclusion}\label{sec5}
	
	In this paper, we propose a novel attribute-guided Transformer network for ZSL (termed TransZero). First, our TransZero employs a feature augmentation encoder to improve the discriminability and transferability of visual features by alleviating the cross-dataset problem and reducing the entangled region feature relationships. Meanwhile, a visual-semantic decoder is introduced to learn the attribute localization for locality-augmented visual feature representations.  Secondly, a visual-semantic embedding network is used to enable effective visual-semantic interaction between the learned locality-augmented visual features and class semantic vectors. Extensive experiments on three popular benchmark datasets demonstrate the superiority of our approach. We believe that our work also facilitates the development of other visual-and-language learning systems, e.g., natural language for visual reasoning.

	\section*{Acknowledgements} This work is partially supported by NSFC~(61772220), Special projects for technological innovation in Hubei Province~(2018ACA135) and Key R\&D Plan of Hubei Province~(2020BAB027).

	%
	\bibliography{mybibfile}
	

\end{document}